\documentclass[sigconf]{acmart}

\renewcommand\footnotetextcopyrightpermission[1]{}

\AtBeginDocument{%
  \providecommand\BibTeX{{%
    \normalfont B\kern-0.5em{\scshape i\kern-0.25em b}\kern-0.8em\TeX}}}

\setcopyright{acmlicensed}
\copyrightyear{2022}
\acmYear{2022}
\acmDOI{10.1145/3503161.3548038}

\acmConference[MM '22]{Proceedings of the 30th ACM International Conference on Multimedia}{October 10--14, 2022}{Lisboa, Portugal}
\acmBooktitle{Proceedings of the 30th ACM International Conference on Multimedia (MM '22), October 10--14, 2022, Lisboa, Portugal}
\acmPrice{15.00}
\acmISBN{978-1-4503-9203-7/22/10}

\acmSubmissionID{1207}

\usepackage{multirow}

\begin{document}

\title{TSRFormer: Table Structure Recognition with Transformers}



\author{Weihong Lin}

\affiliation{%
  \institution{Microsoft Research Aisa}
  \city{Beijing}
  \country{China}
}
\email{weihlin@microsft.com}

\author{Zheng Sun}
\authornote{This work was done when Zheng Sun, Chixiang Ma, Mingze Li and Jiawei Wang were interns in Multi-Modal Interaction Group, Microsoft Research Asia, Beijing, China.}
\affiliation{%
  \institution{University of Chinese Academy of Sciences \& CASIA}
  \city{Beijing}
  \country{China}}
\email{sunzheng2019@gmail.com}

\author{Chixiang Ma}
\authornotemark[1]
\affiliation{%
  \institution{University of Science and Technology of China}
  \city{Hefei}
  \country{China}
}
\email{chixiangma@gmail.com}

\author{Mingze Li}
\authornotemark[1]
\affiliation{%
 \institution{Shanghai Jiao Tong University}
 \city{Shanghai}
 \country{China}}
\email{yagami@sjtu.edu.cn}

\author{Jiawei Wang}
\authornotemark[1]
\affiliation{%
  \institution{University of Science and Technology of China}
  \city{Hefei}
  \country{China}}
\email{wangjiawei@mail.ustc.edu.cn}

\author{Lei Sun}
\affiliation{%
  \institution{Microsoft Research Asia}
  \city{Beijing}
  \country{China}
}
\email{lsun@microsoft.com}

\author{Qiang Huo}
\affiliation{
  \institution{Microsoft Research Asia}
  \city{Beijing}
  \country{China}
}
\email{qianghuo@microsoft.com}

\renewcommand{\shortauthors}{Weihong Lin et al.}

\renewcommand{\thefootnote}{\fnsymbol{footnote}}

\begin{abstract}
We present a new table structure recognition (TSR) approach, called TSRFormer, to robustly recognizing the structures of complex tables with geometrical distortions from various table images. Unlike previous methods, we formulate table separation line prediction as a line regression problem instead of an image segmentation problem and propose a new two-stage DETR based separator prediction approach, dubbed \textbf{Sep}arator \textbf{RE}gression \textbf{TR}ansformer (SepRETR), to predict separation lines from table images directly. To make the two-stage DETR framework work efficiently and effectively for the separation line prediction task, we propose two improvements: 1) A prior-enhanced matching strategy to solve the slow convergence issue of DETR; 2) A new cross attention module to sample features from a high-resolution convolutional feature map directly so that high localization accuracy is achieved with low computational cost. After separation line prediction, a simple relation network based cell merging module is used to recover spanning cells. With these new techniques, our TSRFormer achieves state-of-the-art performance on several benchmark datasets, including SciTSR, PubTabNet and WTW. Furthermore, we have validated the robustness of our approach to tables with complex structures, borderless cells, large blank spaces, empty or spanning cells as well as distorted or even curved shapes on a more challenging real-world in-house dataset.
\end{abstract}




\maketitle
\pagestyle{plain}

\section{Introduction}
Tables offer a means to efficiently represent and communicate structured data in many scenarios like scientific publications, financial statements, invoices, web pages, etc. Due to the trend of digital transformation, automatic table structure recognition (TSR) has become an important research topic in document understanding and attracted the attention of many researchers. TSR aims to recognize the cellular structures of tables from table images by extracting the coordinates of cell boxes and row/column spanning information. This task is very challenging since tables may have complex structures, diverse styles and contents, and become geometrically distorted or even curved during an image capturing process.

Recently, deep learning based TSR methods, e.g., \cite{deepdesrt2017,paliwal2019tablenet,deeptabstr2019,rethinking2019,rethinkinggnn2019,SPLERGE,TabStruct2020,GTE2021,xue2021tgrnet,long2021parsing,FLAG2021,Qiao2021LGPMACT}, have made impressive progress towards recognizing undistorted tables with complex structures and diverse styles. However, these methods except Cycle-CenterNet \cite{long2021parsing} cannot be directly applied to geometrically distorted or even curved tables, which appear often in camera-captured images. Although Cycle-CenterNet \cite{long2021parsing} proposed an effective approach to parsing the structures of distorted bordered tables in wild complex scenes and achieved promising results on their WTW \cite{long2021parsing} dataset, this work didn't take borderless tables into account. Thus, the more challenging problem of recognizing the structures of various geometrically distorted tables still lacks investigation.

In this paper, we propose a new TSR approach, called TSRFormer, to robustly recognizing the structures of both bordered and borderless distorted tables. TSRFormer contains two effective components: 1) A two-stage DETR \cite{deformdetr2021} based separator regression module to directly predict linear and curvilinear row/column separation lines from input table images; 2) A relation network based cell merging module to recover spanning cells by merging adjacent cells generated by intersecting row and column separators. Unlike previous split-and-merge based approaches (e.g., \cite{SPLERGE}), we formulate separation line prediction as a line regression problem instead of an image segmentation problem and propose a new separator prediction approach, dubbed \textbf{Sep}arator \textbf{RE}gression \textbf{TR}ansformer (SepRETR), to predict separation lines from table images directly. In this way, our approach can get rid of heuristic mask-to-line modules and become more robust to distorted tables. Specifically, SepRETR predicts one reference point for each row/column separator first, then takes the features of these reference points as object queries and feeds them into a DETR \cite{detr2020} decoder to regress the coordinates of their corresponding separation lines directly. To make the two-stage DETR framework work efficiently and effectively for the separation line prediction task, we propose two improvements further: 1) A prior-enhanced matching strategy to solve the slow convergence issue of DETR; 2) A new cross attention module to sample features from a high-resolution convolutional feature map directly so that high localization accuracy is achieved with low computational cost. With these new techniques, our TSRFormer has achieved state-of-the-art performance on several public TSR benchmarks, including SciTSR \cite{chi2019complicated}, PubTabNet \cite{zhong2020image} and WTW \cite{long2021parsing}. Furthermore, we have demonstrated the robustness of our approach to tables with complex structures, borderless cells, large blank spaces, empty or spanning cells as well as distorted or even curved shapes on a more challenging real-world in-house dataset.

\begin{figure*}
    \centering
    \includegraphics[width=0.9\textwidth]{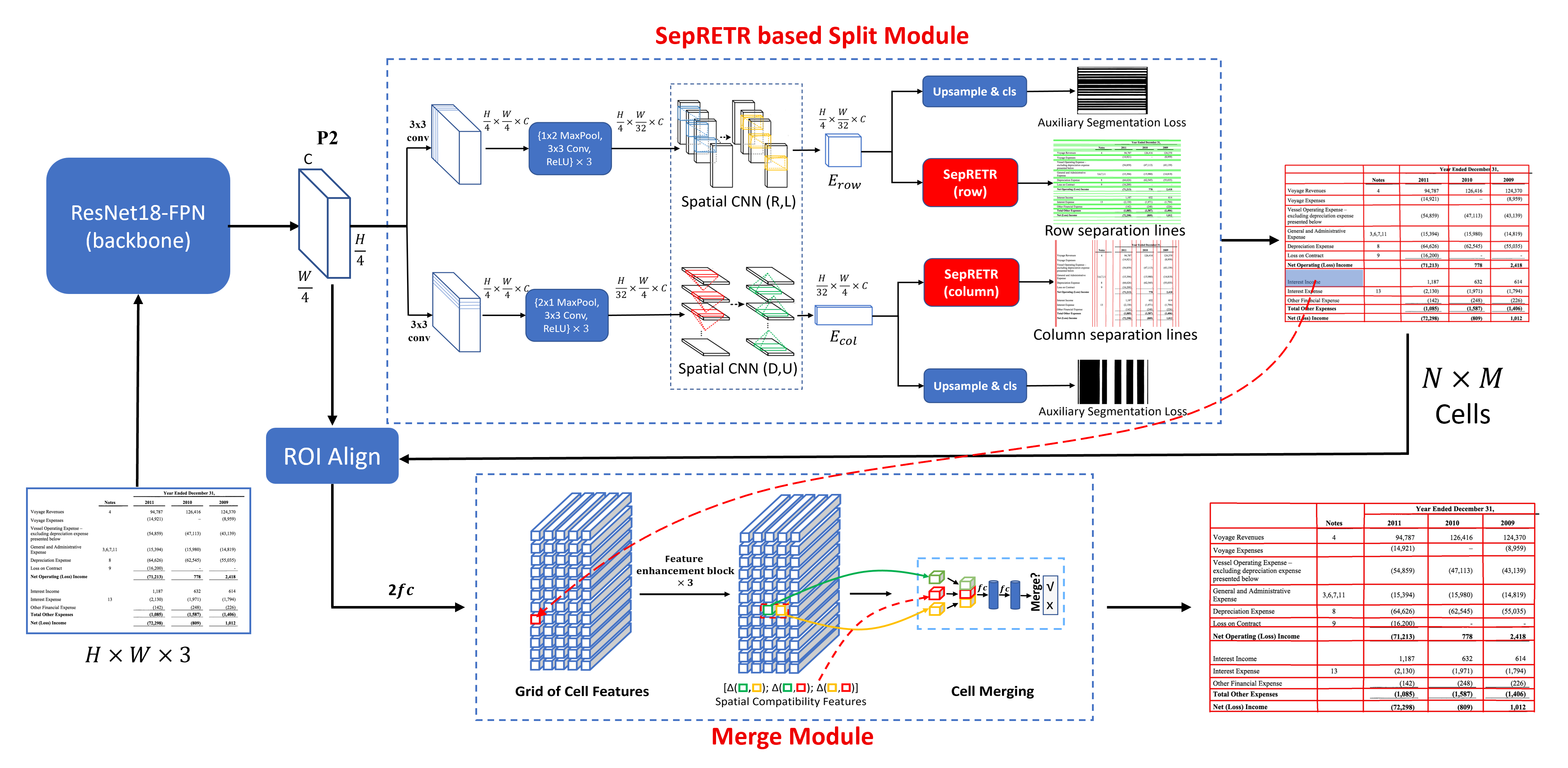}
    \caption{An overview of the proposed TSRFormer.}
    \label{fig-2}
\end{figure*}

\section{Related Work}
\subsection{Table Structure Recognition}
Early TSR methods were mainly based on handcrafted features and heuristic rules (e.g., \cite{laurentini1992identifying,itonori1993table,trecs1998,shigarov2016configurable,rastan2019texus}), so they could only deal with simple table structures or specific data formats, such as PDF files. Later, some statistical machine learning based methods (e.g, \cite{ng1999learning,wang2004table}) were proposed to reduce the dependence on heuristic rules. However, these methods still made strong assumptions about table layouts and relied on handcrafted features, which limited their generalization ability. In recent years, many deep learning based approaches have emerged and outperformed these traditional methods significantly in terms of both accuracy and capability. These approaches can be roughly divided into three categories: row/column extraction based methods, image-to-markup generation based methods and bottom-up methods.

\textbf{Row/column extraction based methods.} These approaches leverage object detection or semantic segmentation methods to detect entire rows and columns first, then intersect them to form a grid of cells. DeepDeSRT \cite{deepdesrt2017} first applied an FCN-based semantic segmentation method \cite{fcn2015} to table structure extraction. TableNet \cite{paliwal2019tablenet} proposed an end-to-end FCN-based model to simultaneously detect tables and recognize table structures. However, these vanilla FCN based TSR methods are not robust to tables containing large blank spaces due to limited receptive fields. To alleviate this problem, methods like \cite{rethinking2019,SPLERGE,khan2019table} tried different context enhancement techniques, e.g., pooling features along rows and columns of pixels on some intermediate feature maps of FCN models or using sequential models like bi-directional gated recurrent unit networks (GRU), to improve row/column segmentation accuracy. Another group of approaches \cite{deeptabstr2019,hashmi2021guided} treated TSR as an object detection problem and used some object detection methods to directly detect the bounding boxes of rows and columns. Among these methods, only SPLERGE \cite{SPLERGE} can deal with spanning cells, which proposed to add a simple cell merging module after a row/column extraction module to recover spanning cells by merging adjacent cells. Later, several works were proposed to further improve the cell merging module. TGRNet \cite{xue2021tgrnet} designed a network to jointly predict the spatial locations and spanning information of table cells. SEM \cite{zhang2022split} fused the features of each cell from both vision and text modalities. Raja et al. \cite{raja2022visual} improved this ``split-and-merge'' paradigm by targeting row, column and cell detection as object detection tasks and forming rectilinear associations through a graph-based formulation for generating row/column spanning information. Different from this two-stage paradigm, Zou et al. \cite{zou2020deep} proposed a one-stage approach to predicting the real row and column separators to handle spanning cells. Although these methods have achieved impressive performance on some previous benchmarks, e.g., \cite{gobel2013icdar,chi2019complicated,zhong2020image}, they cannot handle distorted tables because they rely on an assumption that tables are axis-aligned. Our previous work, RobusTabNet \cite{ma2022robust}, proposed a new split-and-merge based method by incorporating a spatial CNN module \cite{spatialcnn} into an image segmentation based split module to improve its robustness to distorted tables, which makes this new TSR approach able to recognize distorted tables robustly to some extend. However, the performance of this approach is affected by a heuristic mask-to-line module, which struggles with some low-quality separator masks predicted by the split module.

\textbf{Image-to-markup generation based methods.} This type of methods treat TSR as an image-to-markup generation problem and adopt existing image-to-markup models to directly convert each source table image into target presentational markup that fully describes its structure and cell contents. Prior arts tried different image-to-markup models to to convert table images into LaTeX symbols \cite{deng2019challenges,he2021pingan} or HTML sequences \cite{li2020tablebank, zhong2020image}. These methods rely on a large amount of data to train their models and still struggle with big and complex tables \cite{li2020tablebank,zhong2020image}.

\textbf{Bottom-up methods.} Bottom-up methods can be further categorized into two groups. The first group \cite{rethinkinggnn2019,chi2019complicated,li2021gfte,xue2019res2tim} treats words or cell contents as nodes in a graph and uses graph neural networks to predict whether each sampled node pair is in a same cell, row or column. However, the cell contents used by these methods are not directly available when the inputs are table images. To bypass this problem, the second group of methods \cite{GTE2021,prasad2020cascadetabnet,TabStruct2020,li2021adaptive,Qiao2021LGPMACT,FLAG2021} detects the bounding boxes of table cells or cell contents directly and uses different methods to group them into rows and columns. After cell detection, methods like \cite{GTE2021,li2021adaptive,Qiao2021LGPMACT} used heuristic rules to cluster detected cells into rows and columns. CascadeTabNet \cite{prasad2020cascadetabnet} recovered cell relations based on some rules for borderless tables while intersected detected separation lines to extract the grid of bordered tables. TabStruct-Net \cite{TabStruct2020} proposed an end-to-end network to detect cells and predict cell relations jointly. FLAG-Net \cite{FLAG2021} predicted adjacency relationships between detected word bounding boxes rather than cells. However, these approaches fail to handle tables containing a large number of empty cells or distorted/curved tables. Cycle-CenterNet \cite{long2021parsing} detected the vertices and center points of cells simultaneously, and grouped the cells into tabular objects by learning the common vertices. This method can handle curved bordered tables in wild scenes, but does not take borderless tables into account.

\subsection{DETR and Its Variants}
DETR \cite{detr2020} is a novel Transformer-based \cite{transformer2017} object detection algorithm, which introduced the concept of object query and set prediction loss to object detection. These novel attributes make DETR get rid of many hand-designed components in previous CNN-based object detectors like anchor design and non-maximum suppression (NMS). However, DETR has its own issues: 1) Slow training convergence; 2) Unclear physical meaning of object queries; 3) Hard to leverage high-resolution feature maps due to high computational complexity. Deformable DETR \cite{deformdetr2021} proposed several effective techniques to address these issues: 1) Formulating queries as 2D anchor points; 2) Designing a deformable attention module that only attends to certain sampling points around a reference point to efficiently leverage multi-scale feature maps; 3) Proposing a two-stage DETR framework and an iterative bounding box refinement algorithm to further improve accuracy. Inspired by the concept of reference point in Deformable DETR, some follow-up works attempted to address the slow convergence issue by giving spatial priors to the object query. For instance, Conditional DETR \cite{meng2021conditional} divided the cross-attention weights into two parts, i.e., content attention weights and spatial attention weights, and proposed a conditional spatial query to make each cross attention head in each decoder layer focus on a different part of an object. Anchor DETR \cite{wang2021anchor} generated object queries from 2D anchor points directly. DAB-DETR \cite{liu2022dab} proposed to use 4D anchor box coordinates to represent queries and dynamically update boxes in each decoder layer. SMCA \cite{gao2021fast} first predicted a reference 4D box for each query and then directly generated its related spatial cross attention weights with a Gaussian prior in the transformer decoder. Inspired by two-stage Deformable DETR, Efficient DETR \cite{efficientdetr} took top-K scored proposals output from the first dense prediction stage and their encoder features as the reference boxes and object queries, respectively. Different from the above works, TSP \cite{sun2021rethinking} discarded the whole DETR decoder and proposed an encoder-only DETR. Recently, DN-DETR \cite{li2022dn} pointed out that the bipartite matching algorithm used in Hungarian loss is another reason for slow convergence and proposed a denoising based training method to speed up DETR convergence.

\begin{figure*}[t]
    \centering
    \includegraphics[width=0.9\textwidth]{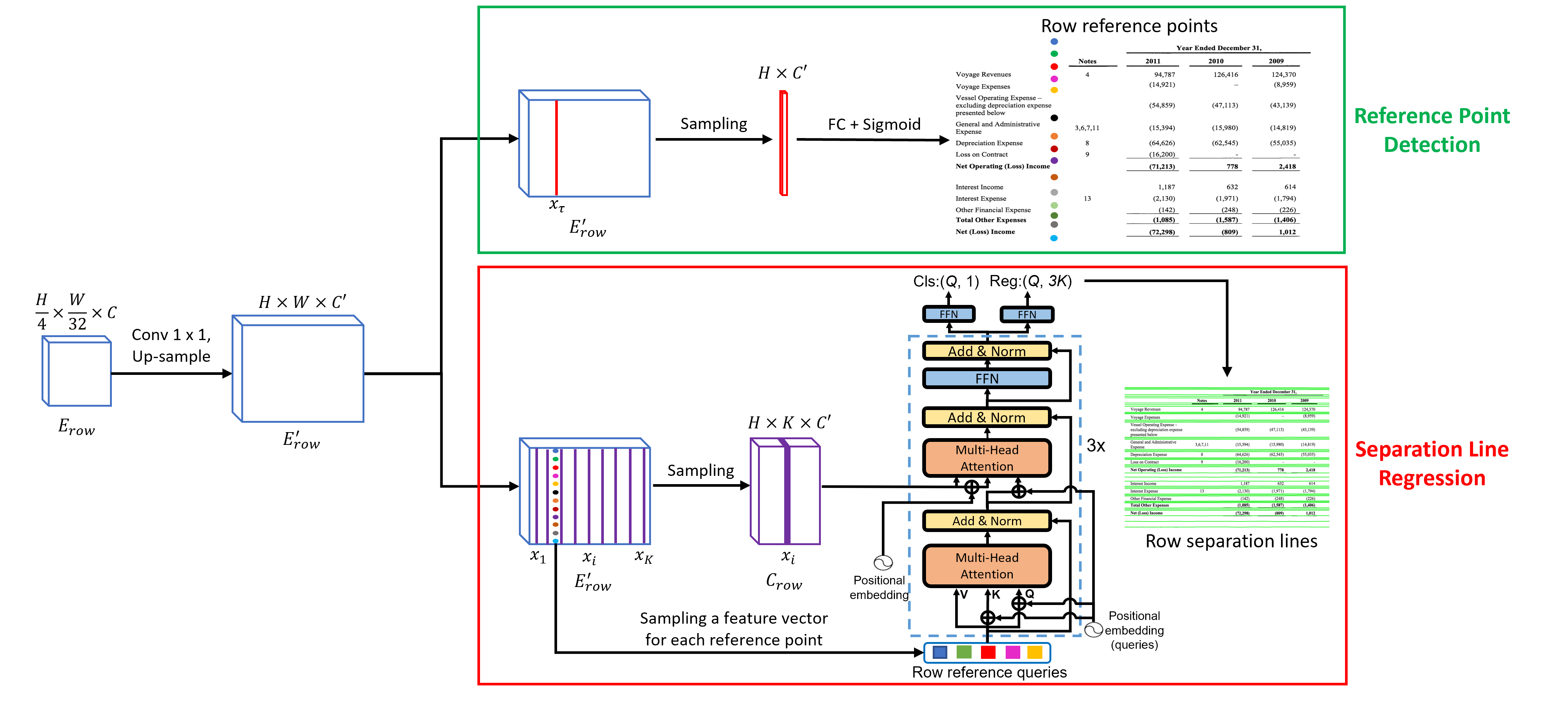}
    \caption{The architecture of our SepRETR for row separation line prediction.}
    \label{fig-3}
\end{figure*}

\section{Methodology}
 As depicted in Fig.~\ref{fig-2}, TSRFormer contains two key components: 1) A SepRETR based split module to predict all row and column separation lines from each input table image; 2) A relation network based cell merging module to recover spanning cells. These two modules are attached to a shared convolutional feature map $P_2$ generated by a ResNet-FPN backbone \cite{resnet,fpn2017}.
 
\subsection{SepRETR based Split Module}
In the split module, two parallel branches are attached to the shared feature map $P_2$ to predict row and column separators, respectively. Each branch comprises three modules: (1) A feature enhancement module to generate a context-enhanced feature map; (2) A SepRETR based separation line prediction module; (3) An auxiliary separation line segmentation module. In subsequent sections, we will take the row separation line prediction branch as an example to introduce the details of these three modules.

\textbf{Feature enhancement.} As shown in Fig.~\ref{fig-2}, we add a $3\times3$ convolutional layer and three repeated down-sampling blocks, each composed of a sequence of a $1\times2$ max-pooling layer, a $3\times3$ convolutional layer and a ReLU activation function, after $P_2$ sequentially to generate a down-sampled feature map $P_2'\in R^{\frac{H}{4}\times\frac{W}{32}\times C}$ first. Then, following \cite{ma2022robust}, two cascaded spatial CNN (SCNN) \cite{spatialcnn} modules are attached to $P_2'$ to enhance its feature representation ability further by propagating contextual information across the whole feature map in rightward and leftward directions. Take the rightward direction as an example, the SCNN module splits $P_2'$ into $\frac{W}{32}$ slices along the width direction and propagates the information slice by slice from left to right. For each slice, it is first sent to a convolutional layer with the kernel size of $9\times1$ and then merged with the next slice by element-wise addition. With the help of SCNN modules, each pixel in the output context-enhanced feature map $E_{row}$ can leverage the structural information from both sides for better representation ability.

\begin{figure}
  \centering
  \includegraphics[width=0.9\linewidth]{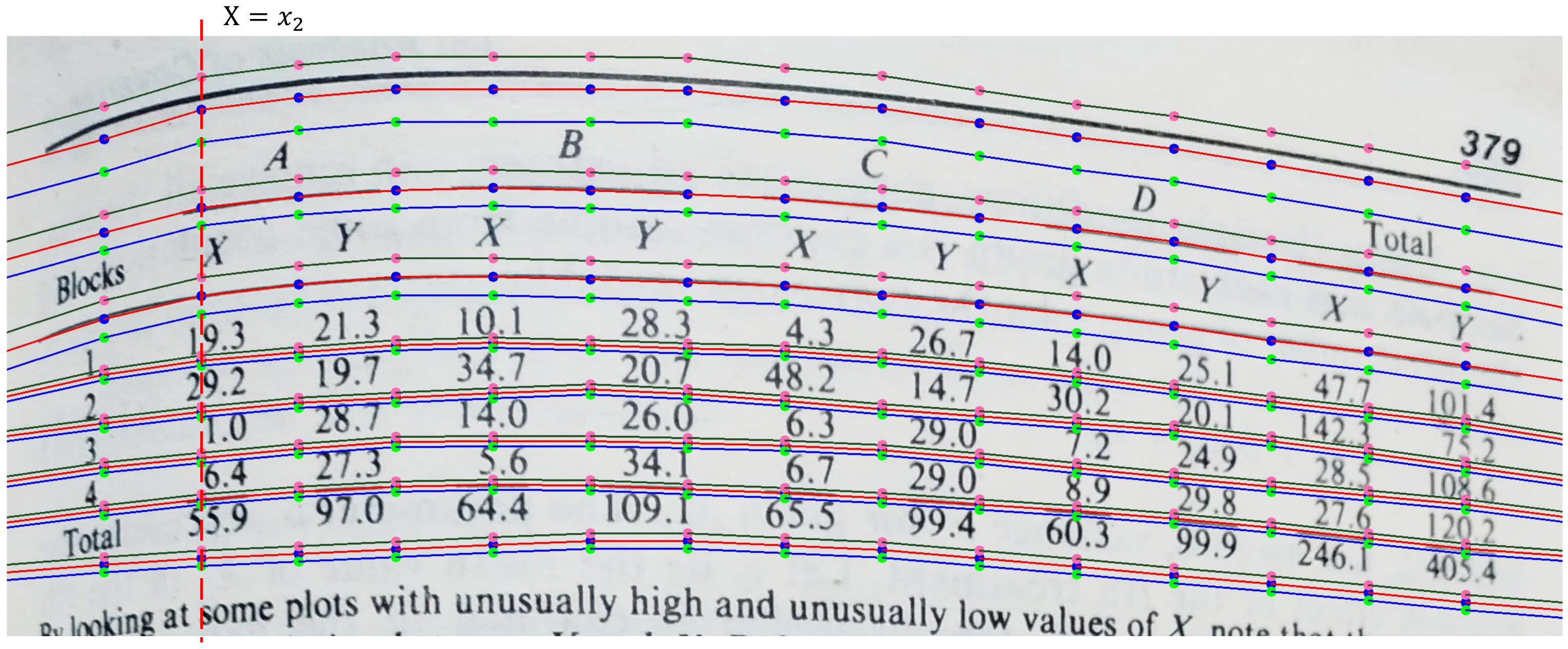}
  \caption{ An example of ground truth row separation lines.
  }
\label{fig-1}
\end{figure}

\textbf{SepRETR based separation line prediction.} As shown in Fig.~\ref{fig-1}, we use three parallel curvilinear lines to represent the top boundary, center line and bottom boundary of each row separator, respectively. Each curvilinear line is represented by $K=15$ points, whose x-coordinates are set to $x_1, x_2, x_3, ..., x_K$, respectively. For each row separator, the y-coordinates of its $3K$ points are predicted by our SepRETR model directly. Here, we set $x_i=\frac{W}{16}\times i$ for the $i^{th}$ x-coordinate. For y-coordinates in the column branch, we only need replace $W$ with $H$. As depicted in Fig.~\ref{fig-3}, our SepRETR contains two modules: a reference point detection module and a DETR decoder for separation line regression. The reference point detection module tries to predict a reference point for each row separator from the enhanced feature map $E_{row}$ first. The features of detected reference points are taken as object queries and fed into a DETR decoder to generate an enhanced embedding for each query. These enhanced query embeddings are then independently decoded into separation line coordinates and class labels by feedforward networks. Both of the two modules are attached to a shared high-resolution feature map $E_{row}'\in R^{H\times\ W\times C'}$, which is generated by adding a $1\times1$ convolutional layer and an up-sampling layer sequentially to $E_{row}$.

\textbf{1) Reference point detection.} This module tries to predict a reference point for each row separator at a fixed position $x_\tau$ along the width direction of the raw image. To this end, each pixel in the $x_\tau^{th}$ column of $E_{row}'$ is fed into a sigmoid classifier to predict a score $p_{i}$ to estimate the probability that a reference point is located at its position $(i,x_\tau)$. Here, we set the hyper-parameter $x_\tau$ as $\lfloor \frac{W}{4} \rfloor$ for row line prediction and $y_\tau$ as $\lfloor \frac{H}{4} \rfloor$ for column line prediction in all experiments. Given the probability of each pixel in the $x_\tau^{th}$ column of $E_{row}'$, we apply non-maximal suppression by using a $7\times1$ max-pooling layer on this column to remove duplicate reference points. After that, top-100 row reference points are selected and further filtered by a score threshold of $0.05$. The remaining row reference points are taken as the object queries of the DETR decoder in the row separation line regression module.

\textbf{2) Separation line regression.} For the sake of efficiency, we don't use transformer encoders to enhance the features output by the CNN backbone. Instead, we concatenate the $x_1^{th}, x_2^{th}, ...$, and $x_K^{th}$ columns of the high-resolution feature map $E_{row}'$ to create a new down-sampled feature map $C_{row}\in R^{H\times K\times C'}$. Then, the features of row reference points extracted from $E_{row}'$ at their positions are treated as object queries and fed into a 3-layer transformer decoder to interact with $C_{row}$ for separation line regression. The positional embedding of position $(x,y)$ is generated by concatenating the sinusoidal embeddings of normalized coordinates $\frac{x}{W}$ and $\frac{y}{H}$, which is the same as in DETR. After enhanced by the transformer decoder, the feature of each query is fed into two feedforward networks for classification and regression, respectively. The ground truth of y-coordinates for row separator regression are normalized to $\frac{y_{gt}}{H}$.

\textbf{Prior-enhanced bipartite matching.} Given a set of predictions and their corresponding ground truth objects from an input image, DETR used the Hungarian algorithm to assign ground-truth labels to the system predictions. However, it is found that the original bipartite matching algorithm in DETR is unstable in the training stage \cite{li2022dn}, i.e., a query could be matched with different objects in a same image in different training epochs, which slows down model convergence significantly. We find that most of the reference points detected in the first stage locate between the top and bottom boundaries of their corresponding row separators consistently in different training epochs, so we leverage this prior information to match each reference point with its closest ground-truth (GT) separator directly. In this way, the matching results will become stable during training. Specifically, we generate a cost matrix by measuring the distance between each reference point and each GT separator. If a reference point is located between the top and bottom boundaries of a GT separator, the cost is set to the distance from this reference point to the GT reference point of this separator. Otherwise, the cost is set to $INF$. Based on this cost matrix, we use the Hungarian algorithm to produce an optimal bipartite matching between reference points and ground truth separators. After getting the optimal matching result, we further remove the pair with cost $INF$ to bypass unreasonable label assignments. The experiments in Sec. 4.4 show that the convergence of our SepRETR becomes much faster with our prior-enhanced bipartite matching strategy.

\textbf{Auxiliary separation line segmentation.} This auxiliary branch aims to predict whether each pixel is located in the region of any separator. We add an up-sampling operation followed by a $1\times1$ convolutional layer and a sigmoid classifier after $E_{row}$ to predict a binary mask $M_{row}\in R^{H\times W\times 1}$ for calculating this auxiliary loss.

\subsection{Relation Network based Cell Merging}
After separation line prediction, we intersect row lines with column lines to generate a grid of cells and use a relation network \cite{rn2017} to recover spanning cells by merging some adjacent cells. As shown in Fig.~\ref{fig-2}, we first use RoI Align algorithm \cite{maskrcnn2017} to extract a $7\times7\times C$ feature map from $P_2$ based on the bounding box of each cell, which is then fed into a two-layer MLP with 512 nodes at each layer to generate a 512-d feature vector. These cell features can be arranged in a grid with $N$ rows and $M$ columns to form a feature map $F_{cell}\in R^{N\times M\times512}$, which is then enhanced by three repeated feature enhancement blocks to obtain wider context information and  fed into a relation network to predict the relationship between adjacent cells. Each feature enhancement block contains three parallel branches with a row-level max-pooling layer, a column-level max-pooling layer and a 3x3 convolutional layer, respectively. The output feature maps of these three branches are concatenated together and convoluted by a $1\times1$ convolutional layer for dimension reduction. In the relation network, for each pair of adjacent cells, we concatenate their features and an 18-d spatial compatibility feature introduced in \cite{rn2017}. A binary classifier is then applied on this feature to predict whether these two cells should be merged or not. The classifier is implemented with a 2-hidden-layer MLP with 512 nodes at each hidden layer and a sigmoid activation function.

\subsection{Loss Function}

The loss functions for training the split module and the cell merging module in TSRFormer are defined in this section. For the split module, we take row separator prediction as an example, and denote the corresponding loss items as $L_{*}^{row}$. Likewise, we can also calculate the losses for column separator prediction, denoted as $L_{*}^{col}$.

\textbf{Reference point detection.} We adopt a variant of focal loss \cite{focal2017} to train the row reference point detection module:
\begin{small}
\begin{equation}
L_{ref}^{row}=-\frac{1}{N_r}\sum_{i=1}^H \left\{
    \begin{array}{ll}
        (1-p_{i})^\alpha log(p_{i}),  &   p^*_{i}=1 \\
        (1-p^*_{i})^\beta p_{i}^\alpha log(1-p_{i}),  &   otherwise
    \end{array}
\right.
\end{equation}
\end{small}

\noindent where $N_r$ is the number of row separation lines, $\alpha$ and $\beta$ are two hyper-parameters set to 2 and 4 respectively as in \cite{cornernet2018}, $p_i$ and $p_i^*$ are the predicted and ground-truth labels for the $i^{th}$ pixel in the $x_\tau^{th}$ column of $E_{row}'$. Here, $p_i^*$ has been augmented with unnormalized Gaussians, which are truncated at the boundary of separators, to reduce the penalty around the ground-truth reference point locations. Specifically, let $(y_k,x_\tau)$ denote the ground-truth reference point for the $k^{th}$ row separator, which is the intersection point of the center line of this row separator and the vertical line $x = x_\tau$. The vertical distance between the top and bottom boundaries of the $k^{th}$ row separator is taken as its thickness, denoted as $w_k$. Then, $p_i^{*}$ can be defined as follows:
\begin{small}
\begin{equation}
p_i^{*}=\left\{
    \begin{array}{ll}
        exp(-\frac{(i-y_k)^2}{2\sigma_k^2}),  &   if~i \in (y_k - \frac{w_k}{2}, y_k + \frac{w_k}{2}) \\
        0,  &   otherwise
    \end{array}
\right.
\end{equation}
\end{small}

\noindent where $\sigma_k=\sqrt{\frac{w_k^2}{2ln(10)}}$ is adaptive to the thickness of the separator to make sure that $p_i^{*}$ within this row separator is no less than 0.1. 

\textbf{Separation line regression.} Let $y=\{(c_i,l_i)|i=1,...,M\}$ denote the set of ground-truth row separators, where $c_i$ and $l_i$ indicate the target class and row separator position respectively, $y^*=\{(c_k^*,l_k^*)|k=1,...,Q\}$ denote the set of predictions. After getting the optimal bipartite matching result $\hat{\sigma}$, the loss of row separation line regression can be calculated as:
\begin{small}
\begin{equation}
    L_{line}^{row}=\sum_{i=1}^Q[L_{cls}(c_i, c_{\hat{\sigma}(i)}^*)+\pmb{1}_{\{c_i\neq \varnothing\}}L_{reg}(l_i,l_{\hat{\sigma}(i)}^*)]
\end{equation}
\end{small}

\noindent where $L_{cls}$ is focal loss and $L_{reg}$ is L1 loss. 

\textbf{Auxiliary segmentation loss.} The auxiliary segmentation loss of row separators is a binary cross-entropy loss:
\begin{small}
\begin{equation}
L_{aux}^{row}=\frac{1}{|S_{row}|}\sum_{(x,y)\in S_{row}} BCE(M_{row}(x,y), M_{row}^*(x,y))
\end{equation}
\end{small}

\noindent where $S_{row}$ denotes the set of sampled pixels from $M_{row}$, $M_{row}(x,y)$ and $M_{row}^*(x,y)$ denote the predicted and ground-truth labels for the pixel $(x, y)$ in $S_{row}$ respectively. $M_{row}^*(x,y)$ is 1 only if this pixel is located within a row separator, otherwise it is 0.

\textbf{Cell merging.} The loss $L_{merge}$ of the cell merging module is a binary cross-entropy loss:
\begin{small}
\begin{equation}
    L_{merge}=\frac{1}{|S_{rel}|}\sum_{i\in S_{rel}} BCE(P_i, P_i^*)
\end{equation}
\end{small}

\noindent where $S_{rel}$ denotes the set of sampled cell pairs,
$P_i$ and $P_i^*$ denote the predicted and ground-truth labels for the $i^{th}$ cell pair, respectively.

\textbf{Overall loss.} All the modules in TSRFormer can be trained jointly. The overall loss function is as follows:
\begin{small}
\begin{align}
    L=\lambda (L_{ref}^{row}+L_{ref}^{col})+L_{aux}^{row}+L_{aux}^{col}+L_{line}^{row}+L_{line}^{col}+L_{merge}
\end{align}
\end{small}

\noindent where $\lambda$ is a control parameter set to 0.2 in our experiments.

\section{Experiments}
\subsection{Datasets and Evaluation Protocols}
 We conduct experiments on three popular public benchmarks, including SciTSR \cite{chi2019complicated}, PubTabNet \cite{zhong2020image} and WTW \cite{long2021parsing}, to verify the effectiveness of the proposed method. Moreover, we also collected a more challenging in-house dataset, which includes many challenging tables with complex structures, borderless cells, large blank spaces, empty or spanning cells as well as distorted or even curved shapes, to demonstrate the superiority of our TSRFormer.

\textbf{SciTSR} \cite{chi2019complicated} contains 12,000 training samples and 3,000 testing samples of axis-aligned tables cropped from scientific literatures. There are also 716 complicated tables selected by authors from the testing set to create a more challenging test subset, called SciTSR-COMP. In this dataset, the cell adjacency relationship metric \cite{gobel2013icdar} is used as the evaluation metric.

\textbf{PubTabNet} \cite{zhong2020image} contains 500,777 training, 9,115 validating, and 9,138 testing images generated by matching the XML and PDF representations of scientific articles. All the tables are axis-aligned. Since the annotations of the testing set are not released, we only report results on the validation set. This work proposed a new Tree-Edit-Distance-based Similarity (TEDS) metric for table recognition task, which can identify both table structure recognition and OCR errors. However, taking OCR errors into account may cause unfair comparison because of different OCR models used by different TSR methods. Some recent works \cite{GTE2021,TabStruct2020,Qiao2021LGPMACT} have proposed a modified TEDS metric named TEDS-Struct to evaluate table structure recognition accuracy only by ignoring OCR errors. We also use this modified metric to evaluate our approach on this dataset.

\textbf{WTW} \cite{long2021parsing} contains 10,970 training images and 3,611 testing images collected from wild complex scenes. This dataset focuses on bordered tabular objects only and contains the annotated information of table id, tabular cell coordinates and  row/column information. We crop table regions from original images for both training and testing, and follow \cite{long2021parsing} to use the cell adjacency relationship (IoU=0.6) \cite{gobel2012methodology} as the evaluation metric of this dataset. 

\begin{figure*}[t]
    \centering
    \includegraphics[width=0.9\linewidth]{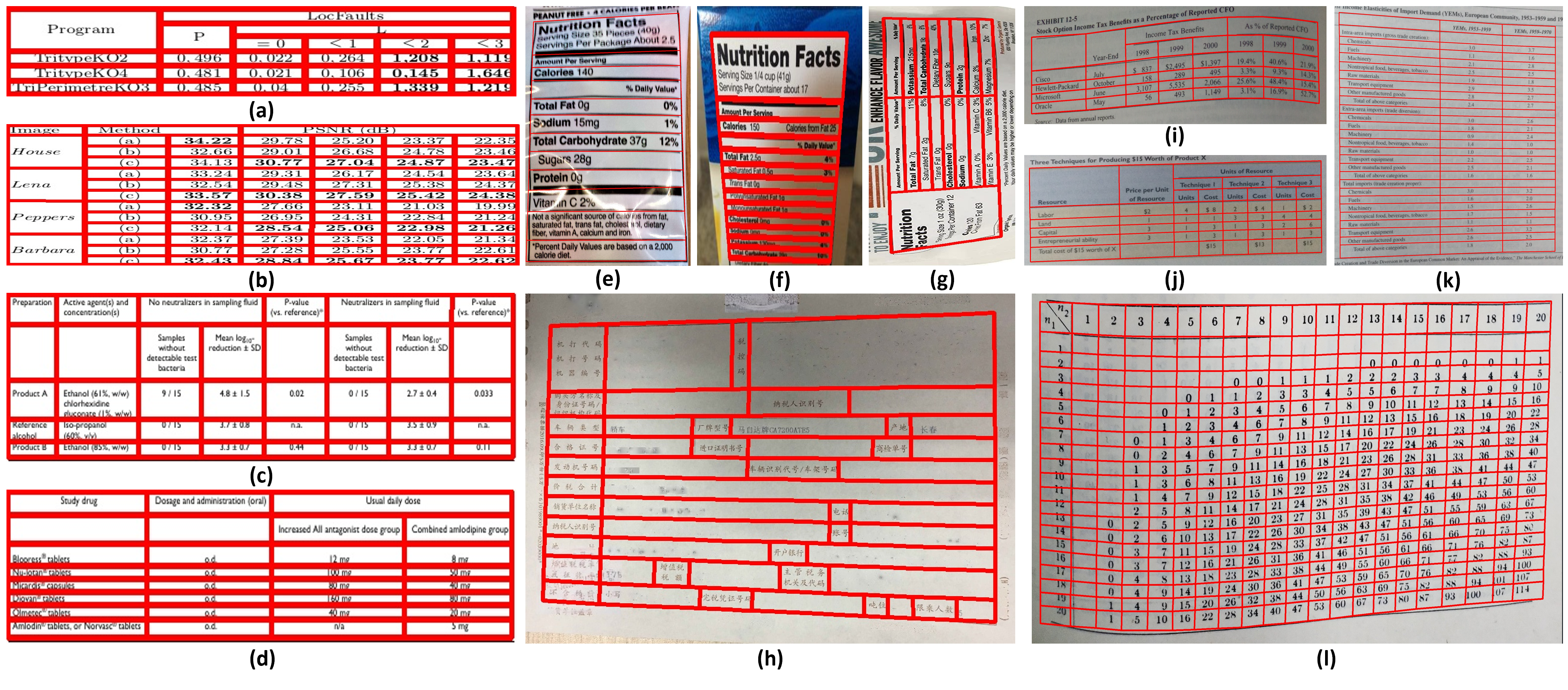}
    \caption{Qualitative results of our approach. (a-b) are from SciTSR, (c-d) are from PubTabNet, (e-h) are from WTW, (i-l) are from the in-house dataset.}
    \label{fig-7}
\end{figure*}

\begin{figure}
    \centering
    \includegraphics[width=0.9\linewidth]{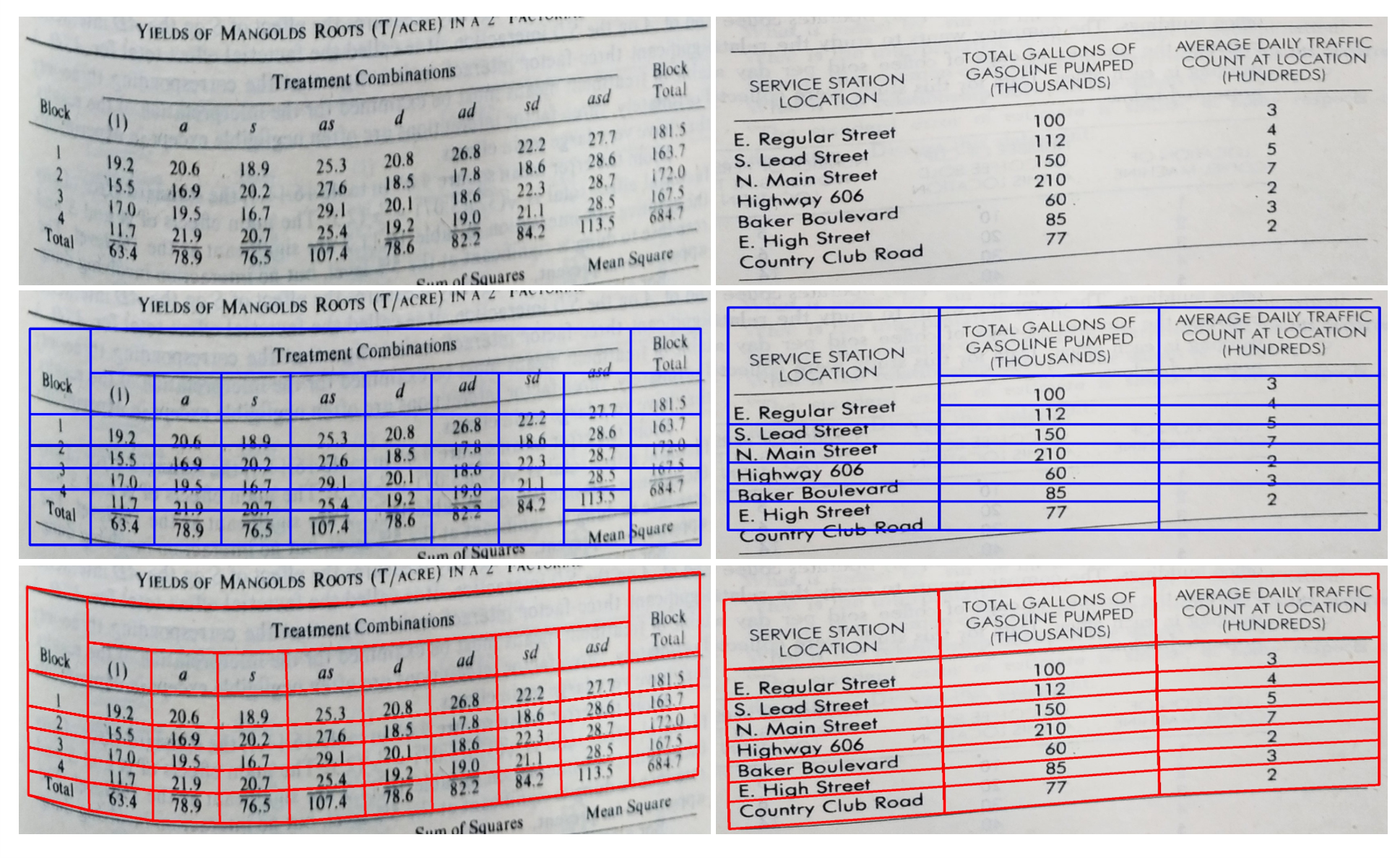}
    \caption{Qualitative results of SPLERGE (blue) and TSRFormer (red) on distorted tables.}
    \label{fig-8}
\end{figure}

\textbf{In-House dataset} contains 40,590 training images and 1,053 testing images, cropped from heterogeneous document images including scientific publications, financial statements, invoices, etc. Most images in this dataset are captured by cameras so tables in these images may be skewed or even curved. Some examples can be found in Fig. \ref{fig-7} and Fig. \ref{fig-8}. The cTDaR TrackB metric \cite{gao2019icdar} is used for evaluation. We use GT text boxes as table contents and report results based on IoU=0.9.

\setlength{\tabcolsep}{4pt}
\begin{table}
\small
\begin{center}
\caption{Results on SciTSR dataset. * denotes the evaluation results without taking empty cells into account.}
\label{table:SciTSR}
\begin{tabular}{ccccccc}
\hline\noalign{\smallskip}
\multirow{2}{*}{Methods} & \multicolumn{3}{c}{SciTSR (\%)} & \multicolumn{3}{c}{SciTSR-COMP (\%)}\\
\cmidrule(lr){2-4} \cmidrule(lr){5-7} & Prec. & Rec. & F1 & Prec. & Rec. & F1\\
\noalign{\smallskip}
\hline
\noalign{\smallskip}
TabStruct-Net \cite{TabStruct2020}  & 92.7 & 91.3 & 92.0 & 90.9 & 88.2 & 89.5\\
GraphTSR \cite{chi2019complicated}  & 95.9 & 94.8 & 95.3 & 96.4 & 94.5 & 95.5\\
LGPMA \cite{Qiao2021LGPMACT}  & 98.2 & 99.3 & 98.8 & 97.3 & 98.7 & 98.0\\
FLAG-Net \cite{FLAG2021}  & \textbf{99.7} & 99.3 & 99.5 & 98.4 & 98.6 & 98.5\\
\hline
TSRFormer  & 99.5 & 99.4 & 99.4 & 99.1 & 98.7 & 98.9\\
TSRFormer* & \textbf{99.7} & \textbf{99.6} & \textbf{99.6} & \textbf{99.4} & \textbf{99.1} & \textbf{99.2} \\
\hline
\end{tabular}
\end{center}
\end{table}
\setlength{\tabcolsep}{1.4pt}

\setlength{\tabcolsep}{4pt}
\begin{table}
\small
\begin{center}
\caption{Results on PubTabNet dataset.}
\label{table:PubTabNet}
\begin{tabular}{cccc}
\hline\noalign{\smallskip}
Methods & Training Dataset & TEDS (\%) & TEDS-Struct (\%)\\
\noalign{\smallskip}
\hline
\noalign{\smallskip}
EDD \cite{zhong2020image}  & PubTabNet & 88.3 & -\\
TableStruct-Net \cite{TabStruct2020}  & SciTSR & - & 90.1\\
GTE \cite{GTE2021}  & PubTabNet & - & 93.0\\
LGPMA \cite{Qiao2021LGPMACT}  & PubTabNet & 94.6 & 96.7\\
FLAG-Net \cite{FLAG2021}  & SciTSR & 95.1 & -\\
\hline
TSRFormer  & PubTabNet & - & \textbf{97.5}\\
\hline
\end{tabular}
\end{center}
\end{table}
\setlength{\tabcolsep}{1.4pt}

\subsection{Implementation Details}
All experiments are implemented in Pytorch v1.6.0 and conducted on a workstation with 8 Nvidia Tesla V100 GPUs. We use ResNet18-FPN as the backbone and set the channel number of $P_2$ to 64 in all experiments. The weights of RestNet-18 are initialized with a pre-trained model for the ImageNet classification task. The models are optimized by AdamW \cite{loshchilov2017decoupled} algorithm with batch size 16. We use a polynomial decay schedule with the power of 0.9 to decay learning rate, and the initial learning rate, betas, epsilon and weight decay are set as
1e-4, (0.9, 0.999), 1e-8 and 5e-4, respectively. Synchronized BatchNorm is applied during training. In SepRETR based split modules, we set the channel number of $E_{row}'$/$E_{col}'$ to 256, and the query dimension, head number and dimension of feedforward networks in transformer decoders to 256, 16 and 1024, respectively.

In the training phase, we randomly rescale the shorter side of table images to a number in \{416, 512, 608, 704, 800\} while keeping the aspect ratio for all datasets except WTW. For WTW, we generate the ground-truth (GT) separation lines by extending the borders of annotated cells and follow \cite{long2021parsing} to resize both sides of each training image to 1024 pixels. Given the GT of separation lines, we follow \cite{ma2022robust} to generate the GT masks of auxiliary segmentation branches in split module and the GT of cell merging module. Then, the center line and two boundaries of each mask will be considered as the GT of regression targets. In each image, a mini-batch of 1024 positive pixels and 1024 negative pixels are randomly sampled for each auxiliary segmentation branch. Furthermore, we sample a mini-batch of 64 hard positive and 64 hard negative cell pairs for the cell merging module. The hard samples are selected with the OHEM \cite{shrivastava2016ohem} alogrithm. During training, we first train the reference point detection and auxiliary segmentation modules jointly for $N$ epochs and then jointly train these two modules and the separation line regression module for $N$ epochs. Finally, the cell merging module is further added and jointly trained for another $N$ epochs. Here, $N$ is set as 12 for PubTabNet and 20 for the other datasets. 

In the testing phase, we rescale the longer side of each image to 1024 while keeping the aspect ratio for SciTSR, PubTabNet and in-house dataset. For WTW, the strategy is the same as in training. 

\begin{table}
\small
\setlength{\tabcolsep}{1.5mm} 
\caption{Results on WTW dataset.}
\label{tab:WTW}
\begin{tabular}{cccc}
    \hline\noalign{\smallskip}
    Methods & Prec. (\%) & Rrec. (\%) & F1-score (\%)\\
    \noalign{\smallskip}
    \hline
    \noalign{\smallskip}
    Cycle-CenterNet \cite{long2021parsing}  & 93.3 & 91.5 & 92.4\\
    \hline
    TSRFormer  & \textbf{93.7} & \textbf{93.2} & \textbf{93.4}\\
    \hline
\end{tabular}
\end{table}

\begin{table}
\small
\setlength{\tabcolsep}{1mm} 
\caption{Comparisons of the enhanced SPLERGE and TSRFormer on different datasets.}
\label{tab:InHouse}
\begin{tabular}{cccccc}
    \hline\noalign{\smallskip}
   Methods & Dataset & Prec. (\%) & Rec. (\%) & F1. (\%) & TEDS-Struct (\%)\\
    \noalign{\smallskip}
    \hline
    \noalign{\smallskip}
    SPLERGE & SciTSR & 99.3 & 98.9 & 99.1 & -\\
    TSRFormer & SciTSR & \textbf{99.5} & \textbf{99.4} & \textbf{99.4} & -\\
    \hline
    SPLERGE & SciTSR-COMP & 98.8 & 98.0 & 98.4 & -\\
    TSRFormer & SciTSR-COMP & \textbf{99.1} & \textbf{98.7} & \textbf{98.9} & -\\
    \hline
    SPLERGE & PubTabNet & - & - & - & 97.1\\
    TSRFormer & PubTabNet & - & - & - & \textbf{97.5}\\
    \hline
    SPLERGE & In-house & 85.4 & 82.3 & 83.8 & -\\
    TSRFormer & In-house & \textbf{95.1} & \textbf{95.3} & \textbf{95.2} & -\\
    \hline
\end{tabular}

\end{table}

\subsection{Comparisons with Prior Arts}
We compare our proposed TSRFormer with several state-of-the-art methods on public SciTSR, PubTabNet and WTW datasets. For SciTSR, since the evaluation tool provided by the authors contains two different settings (consider or ignore empty cells) and some previous works did not explain which one they used, we report the results of both settings. As reported in Table \ref{table:SciTSR}, our approach has achieved state-of-the-art performance on the testing set and the complicated subset, respectively. The excellent result on SciTSR-COMP demonstrates that our method is more robust to complicated tables. On PubTabNet, as shown in Table \ref{table:PubTabNet}, our method has achieved 97.5\% in TEDS-Struct score, which is 0.8\% better than that of LGPMA (the winner of ICDAR 2021 Competition on Scientific Literature Parsing Task B). To verify the effectiveness of our approach on bordered distorted/curved tabular objects in wild scenes, we conduct experiments on WTW dataset and the results in Table \ref{tab:WTW} show that our method is 1.0\% better than Cycle-CenterNet (specially designed for this scenario) in F1-score. 

In order to verify the effectiveness of TSRFormer for more challenging borderless tables, we re-implement another split-and-merge based method SPLERGE \cite{SPLERGE} and compare our approach with it on serveral datasets. For fair comparison, we leverage the same model architecture of TSRFormer and just implement another separation line prediction module which first enhances feature maps by row/column level poolings and then predict axis-aligned separators through classifying pixels in horizantal/vertical slices. As shown in Table \ref{tab:InHouse}, the re-implemented SPLERGE can achieve competitve results on SciTSR and PubTabNet datasets while it is still 11.4\% worse than TSRFormer in F1-score on our challenging in-house dataset. The qualitative results in Fig. \ref{fig-8} and Fig. \ref{fig-7} illustrate that our approach is robust to tables with complex structures, borderless cells, large blank spaces, empty or spanning cells as well as distorted or even curved shapes.

\begin{figure*}
    \centering
    \includegraphics[width=0.9\linewidth]{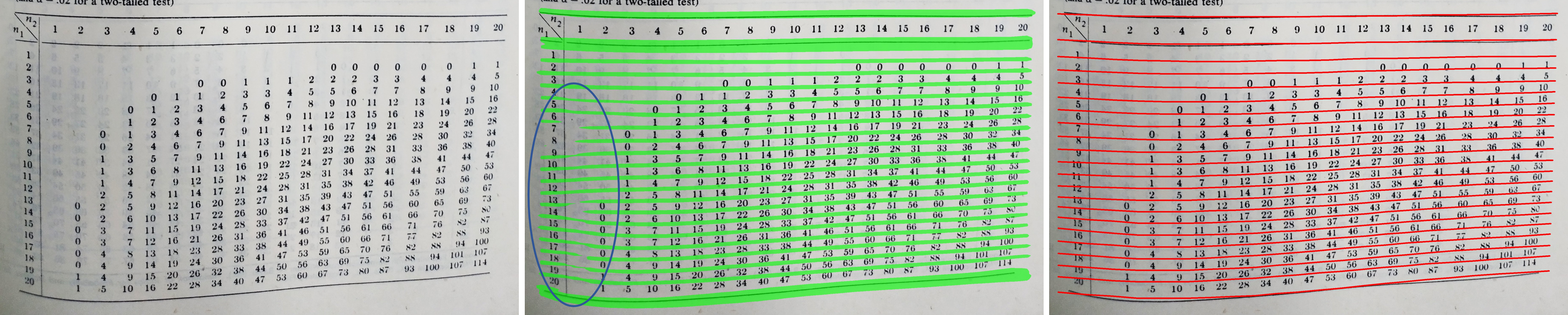}
    \caption{Qualitative results of segmentation based approach with SCNN (middle) and our approach (right) for row separation line prediction on a challenging curved table with borderless cells and large blank spaces.}
    \label{fig-9}
\end{figure*}

\begin{table}
\small
\setlength{\tabcolsep}{1mm} 
\caption{Ablation studies of several modules in TSRFormer.}
\label{tab:ablation1}
\begin{center}
\begin{tabular}{cccccc}
    \hline\noalign{\smallskip}
      & SCNN & Aux-seg. & SepRETR & Cell Merging & F1. (\%)\\
    \noalign{\smallskip}
    \hline
    \noalign{\smallskip}
    \multirow{3}{*}{\shortstack{Segmentation \\ based}}& & \checkmark & & & 83.5\\
    & \checkmark & \checkmark & & & 90.0 \\
    & \checkmark & \checkmark & & \checkmark & 92.3 \\
    \hline
    \multirow{4}{*}{\shortstack{Regression \\ based}}& & & \checkmark & & 88.6\\
    & \checkmark & & \checkmark & & 91.0 \\
    & \checkmark & \checkmark & \checkmark & & \textbf{92.6} \\
    & \checkmark & \checkmark & \checkmark & \checkmark & \textbf{95.2} \\
    \hline
\end{tabular}
\end{center}
\end{table}

\begin{table}
\small
\setlength{\tabcolsep}{1.5mm} 
\caption{Ablation studies of the design of SepRETR.}
\label{tab:ablation2}
\begin{tabular}{cccc}
    \hline\noalign{\smallskip}
    \shortstack{Cross-attention \\ Feature} & \shortstack{Transformer \\ Decoder} & \shortstack{Set \\ Prediction} & F1. (\%)\\
    \noalign{\smallskip}
    \hline
    \noalign{\smallskip}
    --- & & & 90.5 \\
    --- & & \checkmark & 90.7 \\
    $C_{row},C_{col}$ & \checkmark & & 92.2 \\
    $E_{row},E_{col}$ & \checkmark & \checkmark & 92.1 \\ 
    $C_{row},C_{col}$ & \checkmark & \checkmark & \textbf{92.6} \\
    \hline
\end{tabular}
\end{table}

\begin{table}
\small
\setlength{\tabcolsep}{2mm} 
\caption{Effectiveness of prior-enhanced matching strategy.}
\label{tab:ablation3}
\begin{tabular}{ccc}
    \hline\noalign{\smallskip}
    Matching Strategy & \#Epochs & F1. (\%)\\
    \noalign{\smallskip}
    \hline
    \noalign{\smallskip}
    Original in DETR & 20 & 90.1 \\
    Prior-enhanced & 20 & \textbf{92.6} \\
    Original in DETR & 40 & 91.6 \\
    Prior-enhanced & 40 & \textbf{92.8} \\
    \hline
\end{tabular}
\end{table}

\subsection{Ablation Studies}
We conduct a series of experiments to evaluate the effectiveness of different modules in our approach on our in-house dataset.

\textbf{Effectiveness of SepRETR based split module.} To verify the effectiveness of our regression based separator prediction module, we follow RobusTabNet \cite{ma2022robust} to implement another segmentation based split module by removing the SepRETR based separation line regression module and directly using auxiliary segmentation branches for separation line prediction. The heuristic mask-to-line module is also the same as in \cite{ma2022robust}. The results in Table \ref{tab:ablation1} show that our separator regression module is significantly better than the segmentation based split module. Fig. \ref{fig-9} shows some qualitative results. It's very hard for the post-processing module to handle such low-quality masks well. In contrast, our regression based approach is heuristics free and robust to such challenging tables.


\textbf{Ablation studies of the design of SepRETR.} We also conduct the following ablation studies to further examine the contributions of three key components in SepRETR, i.e., the transformer decoder, features used in cross-attention and set prediction. For the experiments without set prediction, we design a heuristic rule for label assignment. If a reference point is located between the two boundaries of a separator, its corresponding query is treated as a positive sample and the regression target is the separator it locates in. Otherwise, the query of this reference point is a negative sample. Since this strategy may assign more than one queries to a separation line, to remove duplicate results, we apply NMS on polygons generated from the two boundaries of each predicted line. As shown in Table \ref{tab:ablation2}, using transformer decoders to help each query leverage both global context and local information can significantly improve the performance of SepRETR based split module. Moreover, the last two rows in Table \ref{tab:ablation2} show that using sampled high-resolution feature map $C_{row}$ and $C_{col}$ can further improve the F1-score by 0.5\%. Although the result without set prediction is good, we find that this approach is very sensitive to some heuristic designs like the rules of label assignment and NMS. On the contrary, training SepRETR with set prediction loss can not only achieve better results, but also get rid of the limitations of such heuristic designs.

\textbf{Effectiveness of prior-enhanced bipartite matching strategy.} We conduct several experiments by training the SepRETR based split module with different matching strategies and epochs. As shown in Table \ref{tab:ablation3}, training the model with the original strategy in DETR by 40 epochs achieves much higher accuracy than training by 20 epochs, which means the split module has not fully converged. In contrast, using the proposed prior-enhanced matching strategy can achieve better results. The small performance gap between models trained with 20 and 40 epochs shows that these two models have converged well, which demonstrates that our prior-enhanced matching strategy can make convergence much faster.

\section{Conclusion}
In this paper, we presented TSRFormer, a new approach for table structure recognition, which contains two effective components: a SepRETR based split module for separation line prediction and a relation network based cell merging module for spanning cell recovery. Compared with previous image segmentation based separation line detection methods, our SepRETR-based separation line regression approach can achieve higher TSR accuracy without relying on heuristic mask-to-line modules. Furthermore, experimental results show that the proposed prior-enhanced bipartite matching strategy can accelerate the convergence speed of two-stage DETR effectively. Consequently, our approach has achieved state-of-the-art performance on three public benchmarks, including SciTSR, PubTabNet and WTW. We have further validated the robustness of our approach to tables with complex structures, borderless cells, large blank spaces, empty or spanning cells as well as distorted or curved shapes on a more challenging real-world in-house dataset.

\clearpage
\bibliographystyle{ACM-Reference-Format}
\bibliography{sample-base}

\end{document}